\documentclass[10pt,conference,letterpaper]{IEEEtran}
\usepackage{times}
\usepackage{graphicx}
\usepackage{latexsym}

\usepackage{helvet}
\usepackage{courier}
\usepackage{amsfonts}
\usepackage{amsmath}
\usepackage{amsthm}
\usepackage[ruled,lined,boxed,commentsnumbered,linesnumbered]{algorithm2e}
\usepackage{xspace}
\usepackage{cite}

\usepackage{graphicx,rotating}

\def\citeauthor#1{\cite #1}
\def\citeyear#1{\cite #1}

\newtheorem{definition}{Definition}

\newcommand{\flow}{\mathsf{flow}}
\newcommand{\jump}{\mathsf{jump}}
\newcommand{\inv}{\mathsf{inv}}
\newcommand{\init}{\mathsf{init}}

\newcommand{\sync}{\mathsf{sync}}

\newcommand{\goal}{\mathsf{goal}}

\newcommand{\lrf}{\mathcal{L}_{\mathbb{R}_{\mathcal{F}}}}

\newcommand{\noop}{\mathsf{noop}}
\newcommand{\trans}{\mathsf{trans}}

\newcommand{\maintain}{\mathsf{maintain}}

\newcommand{\citep}{\cite}
\newcommand{\dreal}{dReal\xspace}

\newcommand{\encodeSMT}{\mathsf{encodeSMT}}
\newcommand{\getTrail}{\mathsf{getTrail}}
\newcommand{\genRun}{\mathsf{genRun}}
\newcommand{\decisions}{\mathsf{decisions}}

\newcommand{\assertLit}{\mathsf{assertLit}}
\newcommand{\retractLit}{\mathsf{retractLit}}
\newcommand{\assertClause}{\mathsf{assertClause}}

\newcommand{\autom}{\mathsf{autom}}
\newcommand{\hide}[1]{}

\newcommand{\mode}{\mathsf{mode}}
\newcommand{\enforce}{\mode}
\newcommand{\hnsolve}{{\tt HNSolve}\xspace}
\newcommand{\etal}{\textit{et al.}\xspace}

\newcommand{\mi}[1]{\mathit{#1}\xspace}

\newcommand{\res}{\mi{res}}
\newcommand{\nil}{\mi{nil}}
\newcommand{\unsat}{\mi{unsat}}
\newcommand{\deltasat}{\ensuremath{\delta}-\ensuremath{\mi{sat}}\xspace}
\newcommand{\consistent}{\mi{consistent}}

\usepackage{xcolor}

\def\und#1{\noindent{\bf #1}:}
\renewcommand{\baselinestretch}{1.0}

\newif\ifblind
\blindfalse

\begin{document}
\title{Instrumenting an SMT Solver to Solve Hybrid Network Reachability Problems}
\ifblind
\author{Paper ID: 6}
\else
\author{\IEEEauthorblockN{Daniel Bryce}\IEEEauthorblockA{SIFT, LLC.,\\ dbryce@sift.net} \and 
\IEEEauthorblockN{Sergiy  Bogomolov}\IEEEauthorblockA{IST Austria,\\ sergiy.bogomlov@ist.ac.at} \and
\IEEEauthorblockN{Alexander Heinz}\IEEEauthorblockA{University of Freiburg, \\ alexander.heinz@saturn.uni-freiburg.de} \and
\IEEEauthorblockN{Christian Schilling}\IEEEauthorblockA{University of Freiburg, \\ schillic@informatik.uni-freiburg.de}}
\fi
\maketitle

\begin{abstract}
PDDL+ planning has its semantics rooted in hybrid automata (HA) and recent
work has shown that it can be modeled as a network of HAs. Addressing
the complexity of nonlinear PDDL+ 
planning as HAs requires both space and time efficient reasoning.
Unfortunately, existing solvers either do not address nonlinear
dynamics or do not natively support networks of automata.  

We  present a new algorithm, called \hnsolve,
which guides the variable selection of the dReal Satisfiability Modulo Theories (SMT)  solver while
reasoning about network encodings of nonlinear PDDL+ planning as HAs.  \hnsolve\ tightly
integrates with dReal by solving a discrete abstraction of the HA
network.  \hnsolve\ finds composite runs on the HA network that
ignore continuous variables, but respect mode jumps and
synchronization labels.  \hnsolve\ admissibly detects dead-ends in the
discrete abstraction, and posts conflict clauses that prune the SMT solver's search.
We evaluate the
benefits of our \hnsolve\ algorithm on PDDL+ benchmark problems and
demonstrate its performance with respect to prior work.
\end{abstract}

\section{Introduction}
Recent 
planners
(\cite{DBLP:conf/aaai/BogomolovMPW14} and \cite{DBLP:conf/aips/BogomolovMMW15}) for PDDL+~\cite{fox2006modelling}
represent actions, processes, events, and state variables as a network of
synchronized hybrid automata (HA), but there are no suitable algorithms for reasoning
about nonlinear change in a network of automata.  We
address nonlinear PDDL+ problems by adapting the dReal Satisfiability Modulo
Theories (SMT) solver
\cite{DBLP:conf/cade/GaoKC13,DBLP:conf/aaai/BryceGMG15}, which has
been previously shown to address nonlinear PDDL+ as a single hybrid
automaton.  It is well-known that reasoning about the explicit
parallel composition of a network of automata as a single automaton is
usually a poor choice because it grows exponentially in the size of
the individual synchronized automata
\cite{DBLP:conf/aaai/CimattiMT12,bu:hal-01055156,bengtsson1998partial}.
We base our HA network SMT encoding  upon prior work
\cite{DBLP:conf/aaai/CimattiMT12} that represents each automaton
independently and adds synchronization constraints.  We show that a direct network encoding results in better scalability
in the number of encoding steps than a single automaton encoding if it
is also matched with an appropriate search strategy.

We extend and reinterpret the approach taken by
Bryce \etal \cite{DBLP:conf/aaai/BryceGMG15} within our proposed \hnsolve\
algorithm. They guide dReal to
systematically select
variable assignments that correspond to discrete-feasible runs of
a hybrid automaton (i.e., ignoring continuous variables).  In this
manner, they force dReal to perform a heuristic depth-first search that considers
all runs through the hybrid automaton.  This led to increased
scalability because dReal did not consider variable assignments
that correspond to discrete-infeasible runs.  A similar technique is
employed by the dReach \cite{DBLP:conf/cade/GaoKC13} and BACH \cite{4689172}
algorithms, but these encode a different SMT instance
for each discrete run and do not benefit from learned conflict
clauses.

\hnsolve, like the work of Bryce \etal \cite{DBLP:conf/aaai/BryceGMG15}, guides
dReal variable selection to construct discrete-feasible runs, but
differs in two main aspects.  First, \hnsolve\ constructs composite
runs for a network of HAs and not a single HA.  Each composite run
represents a feasible sequence of synchronized transitions within each
automaton.  \hnsolve\ also performs a heuristic depth-first search over
the possible composite runs.  The second difference is that
\hnsolve\ also learns conflict clauses that it adds to the SMT
encoding.  While searching the space of discrete-feasible composite runs, \hnsolve
may be unable to find a run that extends the current partial
run.  dReal's current variable assignment encodes a partial run
prefix, and the conflict clause blames the current variable assignment
encoding the prefix.  We find that adding conflict clauses is a
critical new aspect of our approach, and demonstrate that it improves
solver scalability as the number of steps in the encodings increase.  Developing problem-specific
SMT solver algorithms was also recently explored in the
context of program analysis \cite{DBLP:conf/nfm/Schwartz-Narbonne15}.
We hope to advance the understanding of how problem-specific, and arguably
invasive, modifications to SMT solvers can bring about improved performance.

We evaluate the \hnsolve\ algorithm  on several
PDDL+ planning benchmarks encoded as networks of hybrid automata, as
developed in prior work \cite{DBLP:conf/aaai/BogomolovMPW14}.  We compare flat, precompiled encodings of the networks to
our direct encoding of the networks. 
 We show that our encoding outperforms
encodings based upon explicit parallel composition and is competitive
with the state of the art in PDDL+.  We also demonstrate the advantages of
using the \hnsolve\ algorithm to improve performance.
 
\section{Hybrid Automata Background}

We discuss how to represent HA networks in the $\lrf$ language and their
semantics, as follows.

\und{First-order Theories of the Reals}
The $\lrf$ language
represents the first-order signature over the 
reals with the set ${\cal F}$ of computable real functions. 
%  We use the
% {\dreal} SMT solver to find satisfying solutions of, or lack
% thereof these encodings.  In the following, we provide an overview of
% $\lrf$.  

\begin{definition}[$\lrf$-Formulas]
$\lrf$-formulas are first-order formulas over real numbers, whose signature allows an arbitrary collection $\mathcal{F}$ of Type 2 computable real functions~\cite{gao12a}. The syntax is:
\begin{align*}
t& := c\;|\; x \; | \; f(t(\vec x));\\
\varphi& := t(\vec x)> 0 \; | \; t(\vec x)\geq 0 \; | \; \varphi\wedge\varphi
\; | \; \varphi\vee\varphi \; | \; \exists x_i\varphi \; |\; \forall x_i\varphi.
\end{align*}
\end{definition}
A function is Type 2 computable if it can be algorithmically evaluated up to an arbitrary numerical accuracy. 
All common continuous real functions are Type 2 computable. 

% For example, a {\em simplified} portion of the PDDL+ generator domain involves encoding
% the formula:
% \begin{align*}
% &\exists t_k \, \exists \fuellevel_k^0 \, \exists \fuellevel_k^{t_k} . \\
% &\fuellevel_k^{t_k} = \fuellevel_k^0 + \int\limits_0^{t_k} \hspace*{-3pt} \rate_\mi{gen}(s) +
% \rate_\mi{tank}(s) ds \, \wedge \\
% &\quad \forall t \in [0, t_k] .\,
% \fuellevel_k(t) \geq 0
% \end{align*}
% which states that from a relative time zero to time $t_k$ the fuel level just prior to happening $k+1$ is a
% function of the rate the generator consumes fuel and a tank
% replenishes fuel and that the fuel level is no less than zero.

\und{Networks of Hybrid Automata}
We solve HA network reachability problems by
expressing the network in $\lrf$ and then 
unrolling it over $k$ steps and associating a goal region
$\goal(\vec x^t_k)$.  The goal is also an $\lrf$ formula of the form
$\mode_k \wedge \varphi(\vec x^t_k)$ where $\mode_k$ defines the HA
modes to reach and $\varphi(\vec x^t_k)$ defines the variable values to reach.

\begin{definition}[$\lrf$-Representations of Hybrid Automata]\label{lrf-definition}
A hybrid automaton in $\lrf$-representation is a tuple
\begin{align*}
H = \langle &X, Q, \flow, \inv, \jump,\init, {\sf L}\rangle
\end{align*}
where 
\begin{itemize}
\item $X\subseteq \mathbb{R}^n$ for some $n\in \mathbb{N}$,
\item $Q=\{q_1,\ldots,q_m\}$ is a finite set of modes, 
\item $\flow = \{{\flow}_q(\vec x^0, \vec x^t, t): q\in Q\}$ is the set of
ODEs  describing the flow for each mode,  
\item  $\inv = \{\inv_q(\vec x): q\in
Q\}$ is a set of invariants for each  mode, 
\item $\jump = \{\jump_{q\xrightarrow{\ell} q'}(\vec x^t, \vec x^{0'}): q,q'\in
Q\}$, where each element is a transition from mode $q$ to $q'$ using
the synchronization label set $\ell$.  Each formula
$\jump_{q\xrightarrow{\ell} q'}(\vec x^t, \vec x^{0'})$ 
is of the form $\varphi(\vec x^t) \wedge
\psi(\vec x^t, \vec x^{0'})$, where $\varphi(\vec x^t)$ is a
conjunction specifying the guard, $\psi(\vec x^t, \vec x^{0'})$ is
a conjunction specifying the discrete update,
\item $\init = \{\init_q(\vec x): q\in Q\}$ is the set of
initial states, and
\item  ${\sf L}$ is a finite set of
synchronized event labels.
\end{itemize}

% We also assume that for each $q\in
% Q$ there is a ``noop'' transition $\jump_{q\xrightarrow{\{\}} q}(\vec
% x^t, \vec x^{0'})$ where $\vec x^t = \vec x^{0'}$.
\end{definition}

Gao \etal \cite{DBLP:conf/cade/GaoKC13}
 describe how to unroll this encoding
for a single automaton.  The important aspects of the unrolling are to
time stamp each of the continuous variables to denote their value at
the start ($\vec x^0_i$) and end ($\vec x^t_i$) of the mode at step
$i$, as well as the time $t_i$ spent in step $i$. 
% We describe our
% network based encoding in a later section

% The unrolling also
% introduces $\enforce_i = q$ literals to encode the mode transitions
% denoted by jumps. 

%  as well as the times of mode jumps $t_i$ (relative to the
% start of the source mode).  The second line states the initial value
% of each continuous variable, flows, and invariants in the initial
% mode.  The $\enforce$ literal is used to simplify encoding mode
% sequences.  The third and fourth lines encode possible mode
% transitions at each step $i$, as well as how the destination mode
% behaves (i.e., its flows and invariants).  We note that the set of
% possible transitions $h_k(Q)$ will include all transitions at each
% time by default.  The final line encodes the goal that must be
% met in step $k$.  

\begin{definition} (HA Runs)
Each run $\tau$ on $H$ is a series of states 
\begin{align*}
&(t_0, q_0, s(\vec
x_0^0), s(\vec x_0^{t})),\\
&(t_1, q_1,
s(\vec x_1^0), s(\vec x_1^{t})), \\
&\ldots,\\
&(t_k, q_k,
s(\vec x_k^0), s(\vec x_k^{t}))
\end{align*}
 where $t_i$ is the time spent in
step $i$, $q_i$ is a mode, and $s(\vec x_i^0)$ and $s(\vec x_i^{t_i})$
are respective valuations on $\vec x_i^0$ and $\vec x_i^t$ upon entering or leaving
$q_i$.
\end{definition}

The $\init$ set defines what constitutes a legal initial mode $q_0$
and state valuation $s(\vec x_0^0)$.  The $\flow$ and $\inv$ sets
define legal pairs of state valuations $s(\vec x_i^0)$ and $s(\vec
x_i^t)$ and occupancy times $t_i$.  The $\jump$ set defines legal pairs
of modes $q_i$ and $q_{i+1}$, state valuations $s(\vec x_i^t)$ and
$s(\vec x_{i+1}^0)$, and occupancy times $t_i$.

\begin{definition} (Hybrid Automaton Network)
A network ${\cal N} = \{ H_1, \ldots, H_n\}$ of hybrid
automata is a set of hybrid automata.
\end{definition}

\begin{definition} (HA Network Runs)
Each run $\tau$ on ${\cal N}$ is a series of composite states 
\begin{align*}
&(t_0, \vec q_0, s(\vec
x_0^0), s(\vec x_0^{t})),\\
& L_0,\\
& (t_1,\vec q_1,
s(\vec x_1^0), s(\vec x_1^{t})),\\
& L_1,\\
& \ldots,\\
& L_{k-1},\\
&(t_k, \vec q_k,
s(\vec x_k^0), s(\vec x_k^{t}))
\end{align*}
 interleaved with sets of synchronization labels
$L_i$, where each state includes a vector of modes $\vec q_i$ and each
 $\vec x_i^0$ and $\vec x_i^t$  is a valuation on $X = X_1 \cup \ldots \cup X_n$.
\end{definition}

In order for a HA network run to be legal, it must be consistent with
each of the $\init, \flow$, and $\inv$ sets of each individual HA.  Each
label set $L_i$ determines the legal constituent jumps that make up a
composite jump.  In a
composite jump, each automaton either changes modes as defined by its
$\jump$ set, or remains in the same mode. Let $\sync = \ell \cap {\sf
  L}$ be the set of labels a jump must synchronize upon.
 If the composite label is
$L_i$, the automaton must take one of the jumps from the current mode where $\sync \subseteq
L_i$.  If no jump may be taken and ${\sf L} \not\subseteq L_i$, then
the automaton may remain in the same mode.

\section{Hybrid Network Solver}

We describe an extension called the hybrid network solver (\hnsolve) to \dreal's SMT
framework that customizes the problem solver for reachability checking in
networks of hybrid automata.  

\und{ \dreal}
\dreal{} checks whether an $\lrf$ formula is $\delta$-satisfiable (a
decidable problem) by combining a SAT solver
\cite{een2004extensible} with an ICP
solver.\footnote{www.ibex-lib.org} \dreal{} employs the DPLL(T) framework \cite{bruttomesso2010opensmt} for SMT. It first
solves the Boolean constraints to find a satisfying set of literals of
the form $(t(\vec x) \geq 0)$ or $\neg (t(\vec x) \geq 0)$.  This conjunctive set
of literals imposes a set of numeric constraints that are solved using
ICP.  If successful, \dreal{} finishes, and otherwise, the ICP solver
returns a set of literals that explain inconsistency.  The
inconsistent literals become a conflict clause that can be used by the
SAT solver.  If the SAT solver cannot find a satisfying set of
literals, then it returns with an unsatisfiability result.

The ICP solver uses the branch and prune \cite{van1997solving} algorithm to refine
a set of  intervals over the continuous variables (called a box).  Each branch splits the interval
of a single continuous variable, creating two boxes.  Pruning operators propagate the
constraints to shrink the boxes.  
% The primary pruning operator
% enforces hull consistency \cite{granvilliers2006algorithm} of the box.  The constraints that
% involve integrating an ODE are propagated by performing interval-based
% integration with the CAPD library.\footnote{http://capd.ii.uj.edu.pl/}
ICP continues to branch and prune
boxes until it finds a box that is $\delta$-satisfiable or establishes that no
such box exists (i.e., the constraints are inconsistent).  A box is $\delta$-satisfiable 
when for any vector of values $\vec x$ represented by the box, each
constraint $f(\vec x) \geq -\delta$ is satisfied.

\und{Hybrid Network Solver} \hnsolve\ wraps the SAT solver and ICP
solver by suggesting variable assignments and adding conflict clauses.   
Where \dreal\ would normally make decisions without realizing it is
solving a hybrid network reachability problem, \hnsolve\ organizes its
search around potentially viable runs on the network.  To see this,
consider a situation where \dreal\ has made several variable
assignments and decided that $\vec q_0$ and $\vec q_k$ are the initial
and final modes of a run satisfying the $\goal(x_k^t)$.  For the sake
of example, the first automaton $H_1$ does not include jumps that connect
$q_0$ to $q_k$.  A simple shortest path algorithm can recognize that
there is no path from $q_0$ to $q_k$, but \dreal\ will continue to make
assignments unnecessarily and eventually backtrack.

\hnsolve\ solves a discrete abstraction 
\cite{871304,bu:hal-01055156} of the hybrid reachability problem and coordinates the SAT
solver in making assignments corresponding to its solution.  Its
approach is to find a sequence of mode vectors and synchronization labels of
the form 
$$\vec q_0, L_0, \vec q_1, L_1, \ldots, L_{k-1}, \vec q_k$$
that are consistent with the jump transitions and synchronize.
\hnsolve\ then extracts the literal assignments corresponding to this
sequence and guides the SAT solver to realize this run.  If the SAT or ICP
solvers discover an inconsistency, then \hnsolve\ rebuilds the run
from the point to which the SAT solver backtracks.

\hnsolve\ interfaces with the SAT solver through three main methods:
$\getTrail()$, 
$\assertLit()$ and $\assertClause()$.  
The $\getTrail()$ method returns the SAT solver's assignment stack,
including all premises, decision variable assignments, and inferred
variable assignments.
The $\assertLit()$ and $\assertClause()$ methods
assert a literal or clause assignment, respectively, and return
$\unsat$, \deltasat, $\consistent$, or 
{\it backtrack}.  We note that there is no need for a corresponding
$\retractLit()$ method because backtracking is handled by the SAT
solver as part of the calls to $\assertLit()$ and $\assertClause()$.

\hnsolve, Algorithm \ref{alg:hnsolve}, first computes the cost
(described below) to
reach each mode in each automaton in line \ref{alg:hnsolve:cost} and
then encodes the problem in $\lrf$ in line
\ref{alg:hnsolve:encode}.
The
algorithm works in two phases.  In lines \ref{alg:hnsolve:beginpath} -
\ref{alg:hnsolve:endpath}, the solver generates a run suffix ${\cal P}$.
Each run is represented by a set of literals of the form
$\{\enforce_0^1 = q_0^1, \enforce_0^2 = q_0^2, \linebreak[1] \ldots, \sync_0^l,
\ldots, \enforce_k^1 = q_k^1, \enforce_k^2 = q_k^2, \ldots\}$,
defining the modes of each automaton at each step and the
synchronization labels of each transition.  If the current run
represented by the SAT solver's trail cannot be extended, then
$\genRun()$ returns a ``$\nil$'' run.  Lines
\ref{alg:hnsolve:beginclause} - \ref{alg:hnsolve:endclause} create a
conflict clause from the decision variables on the SAT solver's trail
(i.e., negate the corresponding literals) and assert the new clause.
The $\assertClause()$ method returns $\unsat$ when the SAT solver determines
the new clause causes unsatisfiability; otherwise, the SAT solver will
backtrack as part of $\assertClause()$ and we will find a new run on line
\ref{alg:hnsolve:getpath}.

Upon finding a non-$\nil$ run, lines \ref{alg:hnsolve:beginassert} -
\ref{alg:hnsolve:endassert} assert each literal in the
run from the beginning of the run to the end (i.e., following a run
forward to one of the goal modes of each automaton). After
successfully asserting each literal on the run, \hnsolve asserts the
$\nil$ literal, which signals the SAT solver to complete any remaining
assignments. 
 If the run leads to
a \deltasat solution, then the algorithm returns; otherwise, if
asserting a literal is not consistent (i.e., returns {\it backtrack} or
$\unsat$), then the solver attempts to find a new
run on line \ref{alg:hnsolve:getpath}.

\begin{algorithm}[t]
 \SetAlFnt{\tiny}
 \SetAlCapFnt{\small}
 \SetAlCapNameFnt{\small}
\SetAlgoLined
\SetKwFunction{HNSolve}{HNSolve}
\SetKwInOut{Input}{input}\SetKwInOut{Output}{output}
\HNSolve{${\cal N}$, $G$, $k$, $M$}\\
\Input{A network of automata ${\cal N}$, a reachability property $G$, and
  step and delay bound $k$ and $M$.}
$cost \leftarrow {\tt getRunCosts}({\cal N})$\;\label{alg:hnsolve:cost}
$\encodeSMT$(${\cal N}$, $G$, $k$, $M$)\;\label{alg:hnsolve:encode}
\While{true}{
\Repeat {$|{\cal P}| \neq \nil$}{ \label{alg:hnsolve:beginpath}
${\cal P} \leftarrow \genRun({\cal N}, \getTrail(), cost, k)$\;\label{alg:hnsolve:getpath}
 \If{${\cal P} = \nil$}{
   $C \leftarrow \bigvee_{l \in \decisions(\getTrail())} \neg l$\;\label{alg:hnsolve:beginclause} 
   $\res \leftarrow \assertClause$($C$)\;
   \If{$\res = \unsat$}{
     \Return{$\unsat$}
   }\label{alg:hnsolve:endclause} 
 }
}\label{alg:hnsolve:endpath}
 \For{$l$ from 0 upto $|{\cal P}|$}{\label{alg:hnsolve:beginassert}
\uIf{$l < |{\cal P}|$}{
   $\res \leftarrow \assertLit({\cal P}[l])$\;
}
\Else{$\res \leftarrow \assertLit(\nil)$\;
}
   \uIf{$\res$ = \deltasat}{
     \Return{\deltasat}\;
   }
\ElseIf{$\res \neq \consistent$}{
     break\;
   }
}
\label{alg:hnsolve:endassert}
}
\caption{HNSolve algorithm. \label{alg:hnsolve}}
\end{algorithm}

\begin{algorithm}[t]
% \SetAlFnt{\tiny}
% \SetAlCapFnt{\small}
% \SetAlCapNameFnt{\small}
%\SetAlgoLined
%\SetKwFunction{$\genPath$}{$\genPath$}
\SetKwInOut{Input}{input}\SetKwInOut{Output}{output}
$\genRun({\cal N}$, $T$, $cost$, $k$)\\
\Input{A network of automata ${\cal N}$, a stack of literals $T$, a
  cost function $cost$, and
  the step bound $k$.}
${\cal P} \leftarrow []$\;
$S \leftarrow $ dfs(${\cal N}$, $[]$, $cost$, $k$)\;
\eIf{$S$ = fail}{
  {\bf return} nil\;
}{
\For{$j = 0 \ldots |S|-1$}{
  $(q_i \xrightarrow{\ell_i} q_i') \leftarrow S$.get($j$)\;
  $step \leftarrow j/|{\cal N}|$\;
  ${\cal P}$.append(($\mode_{step}^i = q_i'$, $\top$))\;
  \For{$l \in \ell$}{
    ${\cal P}$.append(($\sync_{step}^l$, $\top$))\;
}
}
}
{\bf return} ${\cal P}$\;
\caption{$\genRun$ algorithm. \label{alg:genRun}}
\end{algorithm}

\begin{algorithm}[t]
% \SetAlFnt{\tiny}
% \SetAlCapFnt{\small}
% \SetAlCapNameFnt{\small}
%\SetAlgoLined
\SetKwFunction{dfs}{dfs}
\SetKwInOut{Input}{input}\SetKwInOut{Output}{output}
dfs(${\cal N}$, $S$, $cost$, $k$)\\
\Input{A network of automata ${\cal N}$, a search stack $S$, a mode
  cost function $cost$, and a
  step bound $k$.}
\If{$|S| = |{\cal N}|(k+1)$}{
  {\bf return} $S$\;
  }
$step \leftarrow |S|/ |{\cal N}|$\;
$i \leftarrow |S|\% |{\cal N}|$\;
$succ \leftarrow \{\}$\;
\eIf{$step$ = 0}{
  \For{$\init_{q'}(\vec x) \in \init_i$}{
    $succ \leftarrow succ \; \cup (nil \xrightarrow{\{\}}{q'})$\;
  }
}{
  $(q'' \xrightarrow{\ell'} q) \leftarrow S.$get($|S|-|{\cal N}|$)\;
  \For{$\jump_{q\xrightarrow{L}q'}(\vec x^t, \vec x^0) \in \jump_i$}{
    $succ \leftarrow succ \; \cup (q \xrightarrow{\ell}{q'})$\;
  }
$succ \leftarrow succ \cup (q \xrightarrow{\{\}}{q})$\;
}
$succ \leftarrow $sort(filter($succ$, $i$, ${\cal N}$, $S$, $step$), $cost$)\;
\For{$q\xrightarrow{\ell}{q'} \in succ$}{
  $S$.push($q \xrightarrow{\ell}{q'}$)\;
  \eIf{dfs(${\cal N}$, $S$, $k$) $\not=$ fail}{
    {\bf return} $S$\;
  }{
    $S$.pop()\;
  }
}
  {\bf return} fail\;
\caption{Depth-First Search Algorithm. \label{alg:dfs}}
\end{algorithm}

\begin{algorithm}[t]
% \SetAlFnt{\tiny}
% \SetAlCapFnt{\small}
% \SetAlCapNameFnt{\small}
%\SetAlgoLined
%\SetKwFunction{filter}{filter}
\SetKwInOut{Input}{input}\SetKwInOut{Output}{output}
filter($succ$, $i$, ${\cal N}$, $S$, $step$)\\
\Input{A set of transitions $succ$, an index of the current automaton
  $i$, a network of automata ${\cal N}$, a search stack $S$, and the
  current time step $step$.}
$succ' \leftarrow \{\}$\;
$siblings \leftarrow \{\}$\;
\For{$j = 0 \ldots |S|\%|{\cal N}|$}{
  $siblings \leftarrow siblings \cup  S$.get($|{\cal S}|-(i-j)$)\;
}
\For{$(q_i \xrightarrow{\ell_i} q_i') \in succ$}{
  %Remove negated
  \If{$T$.contains($mode_{step}^i = q_i'$, $\perp$) {\bf or} \\
    \quad $\exists l \in \ell_i .\;  T$.contains($sync_{step}^l$,
    $\perp$) 
% {\bf or } \\
%     \quad $cost[q_i'] > step$
  }{
    {\bf continue}\;
  }
  %Remove costly

  %Remove Nonsync
  $syncs \leftarrow \top$\;
  \For{$ (q_j \xrightarrow{\ell_j} q_j')  \in siblings$}{
    \If{($\ell_j = \{\}$ {\bf and} $q_j = q_j'$ {\bf and} ${\sf L}_i
      \cap {\sf L}_j \cap \ell_i \not= \{\}$) {\bf or }\\
      \quad ($\ell_i = \{\}$ {\bf and} $q_i = q_i'$ {\bf and} ${\sf L}_i
      \cap {\sf L}_j \cap \ell_j \not= \{\}$) {\bf or}\\
      \quad (${\sf L}_i \cap {\sf L}_j \cap \ell_i \not= {\sf L}_i \cap {\sf L}_j
      \cap \ell_j$)}{
      $syncs \leftarrow \perp$\;
      {\bf break}\;
    }
  }
  \If{$syncs$}{
    $succ' \leftarrow succ' \cup (q_i \xrightarrow{\ell_i} q_i')$
  }
}
{\bf return} $succ'$\;
\caption{Filter algorithm. \label{alg:filter}}
\end{algorithm}

The $\genRun()$ method (Algorithm~\ref{alg:genRun}) finds a
run on the network that is consistent with the current SAT solver
assignment $T$. It uses depth-first search (line 3 and Algorithm~\ref{alg:dfs}) to find a search
stack $S$.  The search stack $S$ includes a transition for each
automaton for each step 0 to k-1.  From $S$, the $\genRun$ method extracts the literals
needed to encode the run (lines 7-14).  The literals include mode
choices (line 10) and synchronization labels (lines 11-13).  

The depth-first search (Algorithm~\ref{alg:dfs}) generates the search stack $S$
corresponding to a run on the network.  It selects an initial mode of
$H_1$ at depth 0, an initial mode of $H_2$ at depth 1, and so on.  It
selects a jump from the initial mode of $H_1$ (chosen at depth
0) at depth $|{\cal N}|$, a jump for $H_2$ at depth $|{\cal N}|+1$,
and so on.  Thus, the first ${\cal N}$ levels of the stack correspond
to initial modes, the second $|{\cal N}|$ levels to the zeroth step, and
similarly for later steps.  Lines 8-18
generate the successors, line 19 filters the successors (described below), and lines
20-27 conduct the recursive step of the search.  

Algorithm~\ref{alg:filter} removes successors that are either
inconsistent with the SAT solver's current assignment (lines 8-9) or
do not synchronize with the ``sibling'' jumps previously chosen for the current
step (lines 13-20). The synchronization check involves three cases.
The first and second case are for when either the current automaton or
the sibling will persist its mode (i.e., not synchronize).  We require
that for a mode $q_i$ to persist in automaton ${\cal H}_i$, the
automaton must not be compelled to synchronize with its
siblings at the current time step.  More formally, the mode persistence is allowed if
the public labels ${\sf L}_j \cap \ell_j$ for each  chosen sibling jump $q_j
\xrightarrow{\ell_j} q_j'$ do not intersect with the labels ${\sf
  L}_i$.  The third case checks that two jumps agree on publicly
communicated labels.  The jumps in $succ'$ are those that are possible
in the discrete sense, but may not be possible if we were to consider
their continuous variables in the guards, updates, or mode invariants.

After filtering the possible successors, the depth-first search will
sort the jumps by increasing cost, where cost is defined by the 
successor mode's value in $cost$.  The
cost of each mode is defined by the minimum number of steps from an
initial mode:

\begin{align*}
cost(q) = \left\{
\begin{array}{ll}
0 & : \init_q(\vec x) \in \init_i\\
\min\limits_{\jump_{q' \xrightarrow{\ell} q} \in \jump_i} cost(q') + 1 &: \text{otherwise}
\end{array}\right.
\end{align*}

% \smallskip

% \hnsolve\ asserts literals from the initial modes forward, so
% the current assignment is always a composite run prefix.  The $\genPath()$ method
% extends this prefix by constructing the suffix depth-first.  It
% selects a joint transition among all automata at each depth.  The path
% is discrete-feasible because it respects mode transitions and
% synchronization labels, but ignores continuous variables.  For each
% automaton transition in the joint transition, it  prefers non-noop
% transitions and transitions to modes that have the least
% cost.  The cost of a mode is the minimum number of transitions from an
% initial mode required to reach the mode.  The $\genPath()$ algorithm
% can be thought of as a network generalization of the algorithm defined
% for a single automaton by \cite{DBLP:conf/aaai/BryceGMG15}.

% We use forward search because the PDDL+ problems have a known
% initial state, but only a partial goal state.  By searching forward,
% we avoid directly searching over the possible goal states and only
% consider those that are reachable from the initial state.

In the next section, we detail the $\lrf$ encoding that we use to express
the network HA reachability problem.  We follow with a section
describing  how we experiment with \hnsolve\ by omitting it
entirely (i.e., use dReal alone), omitting lines 6-12 of Algorithm~\ref{alg:hnsolve} to avoid
learning conflict clauses, or using it in its entirety.  We note that
omitting conflict clauses results in an algorithm similar to that
described by Bryce \etal \cite{DBLP:conf/aaai/BryceGMG15}, aside from our generalization
to a network of automata.

\section{Network Encoding} \label{sec:otf}

We encode the parallel
composition of a network of hybrid automata implicitly, as follows.
We encode the mode at step $i$ of each automaton with literals of
  the form $\enforce_i^1 = q_1, \ldots, \enforce_i^m = q_m$. We constrain the possible composite jumps with
 synchronized jump constraints, and noops.  With these constraints,
 we avoid pre-computing all possible $O(2^{|\jump|})$
parallel jumps per step. Instead, we encode $O(|\jump|m)$ synchronized
jump constraints and $O(|Q|m)$ noops.
% The explicit encoding materializes each possible parallel jump and can
% encode each uneffected variable $x$ with persistence constraints of the form
% $x_i^t = x_{i+1}^0$. 
To determine which jumps must synchronize, we
introduce literals for each label and constrain their values with the
appropriate jumps.  Noop (stutter) clauses encode cases where an automaton does not synchronize any
of its transitions, and its mode persists.

\medskip

We define the implicit parallel composition for a $k$-step $M$-delay
reachability problem as the conjunction of clauses describing each
individual automaton, and the goal:

\begin{align*}
& \exists^X \vec x^0_0\,  \exists^X\vec x_0^t\ldots \exists^X \vec
x^0_k\, \exists^X\vec x_k^t\\
& \exists^{[0,M]}t_0\ldots \exists^{[0,M]}t_k
 %\exists^{[0,1]} \vec
 % \gamma_0\ldots\, \exists^{[0,1]} \vec \gamma_k\, 
% \exists \vec \sync_0
%   \ldots \exists \vec \sync_{k-1}
.
\\
& %\toplevel({\cal N}, k)   \wedge 
\left(\bigwedge_{j = 1}^n \autom(H_j, k)\right)
  \wedge \goal(\vec x^t_k) \wedge \left(\bigwedge_{i=0}^{k-1}
  \bigvee_{l\in {\sf L}_1 \cup \ldots \cup {\sf L}_n} \sync_i^l \right)
\end{align*}
The $\vec x_i^0$ and $\vec x_i^t$ variables denote the values of
continuous variables at the start and end of step $i$.  The $t_i$
variables denote the duration of step $i$.  The encoding ensures that
each automaton behaves appropriately, the goal is satisfied, and at least one non-noop transition occurs in each step.
The $\autom(H_j, k)$ clauses define
the behavior of each automaton $H_j$ as:
\begin{align*}
&\hspace*{-1cm}\init_{j}(\vec x^0_0) \wedge \bigwedge_{i=0}^{k} 
\maintain_j(i) \wedge  \\
\bigg{[} \bigwedge_{i=0}^{k-1} & \left(\bigvee_{q\in Q_j} \noop(q,
    i)\right)\vee \\
&\bigvee_{\substack{\jump_{q\xrightarrow{\ell} q'}(\vec x_{i}^{t},
\vec x^{0}_{i+1}) \in \jump_j}}\trans_j(\jump_{q\xrightarrow{\ell} q'}(\vec x_{i}^{t},
\vec x^{0}_{i+1}), i) \bigg{]}
\end{align*}
which constrain the initial state, the continuous change in each mode
at each step, and the transitions between steps.

The clause $\init_{j}(\vec x^0_0)$ constrains the initial values of the
variables and the initial mode.  It defines:
\begin{align*}
\bigvee_{\init_q(\vec x)\in \init_j}
\init_{q}(\vec x^0_0) \wedge (\enforce_0^j = q) 
\end{align*}
to constrain the assignments to $\vec x_0^0$ and the initial mode.

The $\maintain_j(i) $
clause defines how the flows and invariants
of the automaton $H_j$ govern continuous change:
\begin{align*}
&\flow_j(\vec x^0_i, \vec x_i^t, t_i) \, \wedge\\
&\forall^{[0,t_i]}t\ \forall^X\vec x_{i}\;(\flow_j(\vec
x^0_i, \vec x_i, t) \rightarrow \inv_j(\vec x_i))
\end{align*}
where we note that the nested universal quantifiers ensure
that the invariant holds for the entire time the mode is occupied.
The nested quantifiers are a unique aspect of our encoding that
enables us to reason about nonlinear change \cite{gao2013satisfiability}.

The $\flow_j(\vec
x^0_{i}, \vec x_{i}^t, t_i)$ clause defines
\begin{align*}
\bigwedge_{q \in Q_j} (\mode_i^j = q) \rightarrow \flow_q(\vec
x^0_{i}, \vec x_{i}^t, t_i)
\end{align*}
The $\inv_j(\vec x_{i}) $ clause enables the invariants of
the active modes by defining:
\begin{align*}
\bigwedge_{q \in Q_j} (\enforce_i^j = q)  \rightarrow
\inv_{q}(\vec x_{i})
\end{align*}

Noop clauses $\noop(q, i)$ model asynchronous behavior where
the automaton  does not synchronize, and define:
\begin{align*}
\left(\bigwedge_{l \in {\sf L}_j} \neg \sync^l_i\right)  \wedge 
%\left( \bigwedge_{e \in E^j(x')}  \neg e \right)\wedge
  (\enforce_i^j = q) \wedge  (\enforce_{i+1}^j = q) 
\end{align*}
% where $E^j(x')$ denotes the set of all literals constraining $x'$ that
% appear in a clause in $\jump_j$ and the $\sync$ literals.  Omitting the $\noop(q, i)$ clauses
% will force the network to be synchronous.

% The clause $\trans_j(i)$ defines how the automaton transitions from
% step $i$ to $i+1$. It defines:

% \begin{align*}
% \bigvee_{\substack{\jump_{q\xrightarrow{\ell} q'}(\vec x_{i}^{t},
% \vec x^{0}_{i+1}) \in \jump_j}}\trans_j(\jump_{q\xrightarrow{\ell} q'}(\vec x_{i}^{t},
% \vec x^{0}_{i+1}), i) 
% \end{align*}

Jump transition clauses $\trans_j(\jump_{q\xrightarrow{\ell} q'}(\vec x_{i}^{t},
\vec x^{0}_{i+1}), i)$ define how each jump
must synchronize and constrain the variables and
modes:

\begin{align*}
&\left(\bigwedge_{l \in \ell\cap {\sf L}_j} \sync^l_i \wedge \bigwedge_{l \in {\sf L}_j
  \backslash \ell}  \neg  \sync^l_i \right)  \wedge \jump_{q\xrightarrow{\ell} q'}(\vec x_{i}^{t},
\vec x^{0}_{i+1}) \, \wedge\\
&(\enforce_i^j = q) \wedge (\enforce_{i+1}^j = q')
\end{align*}

\begin{table*}[t]
\centering
\begin{tabular}{|@{ }r@{ }|@{ }r@{ }|@{ }r@{ }||@{ }r@{ }|@{ }r@{ }|@{ }r@{ }||@{ }r@{ }|@{ }r@{ }|@{ }r@{ }||@{ }r@{ }|@{ }r@{ }|@{ }r@{ }|}
\hline
k	&	Dom	&	Inst	&	F	&	F+H	&	F+H+L	&	C	&	C+H	&	C+H+L	&	N	&	N+H	&	N+H+L	\\
\hline\hline																							
3 (3)	&	Gen	&	0	&	0.22	&	0.16	&	0.17	&	1.52	&	0.11	&	0.12	&	0.30	&	0.13	&	0.13	\\
7 (5)	&	Gen	&	1	&	-	&	1.66	&	1.15	&	-	&	3.45	&	3.41	&	1.39	&	0.75	&	0.77	\\
11 (7)	&	Gen	&	2	&	-	&	-	&	-	&	-	&	735.00	&	738.10	&	20.69	&	3.62	&	3.62	\\
15 (9)	&	Gen	&	3	&	-	&	-	&	-	&	-	&	-	&	-	&	-	&	12.23	&	12.77	\\
19 (9)	&	Gen	&	4	&	-	&	-	&	-	&	-	&	-	&	-	&	-	&	37.77	&	36.05	\\
23 (13)	&	Gen	&	5	&	-	&	-	&	-	&	-	&	-	&	-	&	-	&	100.21	&	92.26	\\
27 (17)	&	Gen	&	6	&	-	&	-	&	-	&	-	&	-	&	-	&	-	&	511.00	&	360.63	\\
31 (19)	&	Gen	&	7	&	-	&	-	&	-	&	-	&	-	&	-	&	-	&	561.06	&	482.03	\\
\hline																							
6	&	Car1	&	1	&	0.85	&	0.84	&	0.89	&	9.77	&	1.21	&	0.98	&	0.9	&	1.33	&	1.32	\\
5	&	Car1	&	2	&	1.59	&	0.75	&	0.74	&	13.18	&	0.97	&	0.9	&	0.84	&	1.09	&	1.05	\\
5	&	Car1	&	3	&	0.99	&	0.72	&	0.72	&	44.64	&	1.22	&	1.15	&	1.82	&	1.56	&	1.59	\\
5	&	Car1	&	4	&	1.63	&	0.86	&	0.89	&	83.55	&	1.63	&	1.49	&	1.79	&	2.22	&	2.43	\\
5	&	Car1	&	5	&	7.41	&	1.39	&	1.43	&	229.54	&	2.09	&	1.93	&	2.62	&	3.34	&	3.95	\\
5	&	Car1	&	6	&	10.01	&	1.64	&	1.69	&	448.62	&	2.82	&	2.56	&	7.05	&	5.12	&	6.11	\\
5	&	Car1	&	7	&	9.98	&	1.92	&	1.98	&	-	&	3.94	&	3.57	&	7.57	&	7.93	&	9.77	\\
5	&	Car1	&	8	&	10.69	&	1.67	&	1.68	&	-	&	6.78	&	6.26	&	14.98	&	14.87	&	18.33	\\
5	&	Car1	&	9	&	18.65	&	1.96	&	1.94	&	-	&	7.69	&	7.34	&	23.55	&	21.41	&	24.13	\\
5	&	Car1	&	10	&	46.74	&	2.62	&	2.51	&	-	&	12.42	&	10.15	&	40.28	&	34.66	&	40.01	\\
\hline																							
12	&	Car2	&	1	&	-	&	25.02	&	78.63	&	-	&	-	&	-	&	37.45	&	236.39	&	9.87	\\
10	&	Car2	&	2	&	-	&	336.86	&	330.61	&	-	&	-	&	-	&	-	&	-	&	11.84	\\
10	&	Car2	&	3	&	-	&		&	-	&	-	&	-	&	-	&	-	&	-	&	24.98	\\
10	&	Car2	&	4	&	-	&		&	-	&	-	&	-	&	-	&	-	&	-	&	20.04	\\
10	&	Car2	&	5	&	-	&		&	-	&	-	&	-	&	-	&	-	&	-	&	72.35	\\
10	&	Car2	&	6	&	-	&		&	-	&	-	&	-	&	-	&	-	&	-	&	119.69	\\
10	&	Car2	&	7	&	-	&		&	-	&	-	&	-	&	-	&	-	&	-	&	194.38	\\
10	&	Car2	&	8	&	-	&		&	-	&	-	&	-	&	-	&	-	&	-	&	408.78	\\
10	&	Car2	&	9	&	-	&		&	-	&	-	&	-	&	-	&	-	&	-	&	222.63	\\
10	&	Car2	&	10	&	-	&		&	-	&	-	&	-	&	-	&	-	&	-	&	328.82	\\\hline
\end{tabular}
 \caption{\label{tab:internal-linear}Runtime (s) on linear instances. ``-'' indicates a timeout.}
\end{table*}

\section{Empirical Evaluation}

Our evaluation studies the effectiveness of our network encoding and
\hnsolve\ algorithm.  Specifically, we compare three configurations of
our solver on several hybrid automaton encodings of PDDL+ problems.
The configurations include an unmodified dReach/dReal solver, the
addition of the \hnsolve\ algorithm without clause learning, and \hnsolve
with clause learning.  We evaluate the configurations on single automaton encodings \cite{DBLP:conf/aaai/BryceGMG15}, and networks of automata
encodings based on that of Bogomolov \etal \cite{DBLP:conf/aaai/BogomolovMPW14}.  With the
network encodings, we either take the parallel composition and encode
a single automaton, or encode the network as described in the
previous section.  

We note that the encodings used in prior work differ in whether they
include a ``lock'' for the actions. Bryce \etal
\cite{DBLP:conf/aaai/BryceGMG15} hand-encode a single automaton
for each problem that ensures no two actions can occur at the same
time.  In a network of automata, where each action is represented by
its own automaton, Bogomolov \etal \cite{DBLP:conf/aaai/BogomolovMPW14} ensure that no two actions
occur, start, or end simultaneously by introducing a lock automaton.
The network of automata models each action so that it must acquire and release the
lock when it occurs (atomic actions) or starts/ends (durative
actions).  This causes the network encoding to require twice the number
of encoding steps than the single automaton encoding used by Bryce
\etal \cite{DBLP:conf/aaai/BryceGMG15}.  We notice that the two step
lock is only necessary when enforcing $\epsilon$-separation of the
actions.  Since Bryce
\etal  \cite{DBLP:conf/aaai/BryceGMG15} do not model
$\epsilon$-separation we can match their required number of encoding steps with
the network of automata by using a single lock transition that synchronizes with each action.

\begin{table*}[t]
\centering
\begin{tabular}{|@{ }r@{ }|@{ }r@{ }|@{ }r@{ }||@{ }r@{ }|@{ }r@{ }|@{ }r@{ }||@{ }r@{ }|@{ }r@{ }|@{ }r@{ }||@{ }r@{ }|@{ }r@{ }|@{ }r@{ }|}
\hline
k	&	Dom	&	Inst	&	F	&	F+H	&	F+H+L	&	C	&	C+H	&	C+H+L	&	N	&	N+H	&	N+H+L	\\
\hline\hline																							
3 (3)	&	Gen	&	0	&	0.15	&	0.16	&	0.14	&	1.40	&	0.14	&	0.15	&	0.49	&	0.49	&	0.18	\\
7 (5)	&	Gen	&	1	&	0.76	&	1.66	&	1.77	&	-	&	4.21	&	4.22	&	1.74	&	1.16	&	1.14	\\
11 (7)	&	Gen	&	2	&	26.36	&	-	&	-	&	-	&	879.36	&	876.92	&	173.78	&	6.08	&	5.78	\\
15 (9)	&	Gen	&	3	&	-	&	-	&	-	&	-	&	-	&		&	310.70	&	22.91	&	21.77	\\
19 (9)	&	Gen	&	4	&	-	&	-	&	-	&	-	&	-	&		&	-	&	69.16	&	66.89	\\
23 (13)	&	Gen	&	5	&	-	&	-	&	-	&	-	&	-	&	-	&	-	&	320.21	&	316.45	\\
27 (17)	&	Gen	&	6	&	-	&	-	&	-	&	-	&	-	&	-	&	-	&	530.77	&	1044.27	\\
\hline																							
6	&	Car1	&	1	&	5.49	&	1.63	&	1.54	&	-	&	-	&	-	&	3.46	&	2.81	&	23.48	\\
5	&	Car1	&	2	&	3.45	&	1.46	&	1.38	&	-	&	-	&	-	&	2.26	&	2.19	&	22.35	\\
5	&	Car1	&	3	&	8.06	&	1.48	&	1.44	&	-	&	-	&	-	&	6.15	&	3.87	&	43.57	\\
5	&	Car1	&	4	&	4.73	&	1.51	&	1.52	&	-	&	-	&	-	&	7.88	&	7.27	&	81.93	\\
5	&	Car1	&	5	&	5.25	&	1.53	&	1.49	&	-	&	-	&	-	&	10.91	&	9.82	&	145.64	\\
5	&	Car1	&	6	&	6.42	&	1.47	&	1.50	&	-	&	-	&	-	&	19.96	&	17.05	&	251.90	\\
5	&	Car1	&	7	&	7.02	&	1.45	&	1.53	&	-	&	-	&	-	&	42.53	&	29.73	&	465.09	\\
5	&	Car1	&	8	&	9.79	&	1.44	&	1.53	&	-	&	-	&	-	&	76.78	&	45.12	&	216.67	\\
5	&	Car1	&	9	&	10.23	&	1.93	&	2.05	&	-	&	-	&	-	&	143.52	&	76.76	&	356.15	\\
5	&	Car1	&	10	&	12.45	&	1.92	&	2.05	&	-	&	-	&	-	&	221.08	&	121.31	&	498.48	\\
\hline																							
12	&	Car2	&	1	&	-	&	313.77	&	219.1	&	-	&	-	&	-	&	-	&	-	&	24.02	\\
10	&	Car2	&	2	&	-	&	966.82	&	342.12	&	-	&	-	&	-	&	-	&	-	&	23.17	\\
10	&	Car2	&	3	&	-	&	-	&	-	&	-	&	-	&	-	&	-	&	-	&	46.51	\\
10	&	Car2	&	4	&	-	&	-	&	-	&	-	&	-	&	-	&	-	&	-	&	85.67	\\
10	&	Car2	&	5	&	-	&	-	&	-	&	-	&	-	&	-	&	-	&	-	&	146.43	\\
10	&	Car2	&	6	&	-	&	-	&	-	&	-	&	-	&	-	&	-	&	-	&	246.70	\\
10	&	Car2	&	7	&	-	&	-	&	-	&	-	&	-	&	-	&	-	&	-	&	448.86	\\
10	&	Car2	&	8	&	-	&	-	&	-	&	-	&	-	&	-	&	-	&	-	&	217.37	\\
10	&	Car2	&	9	&	-	&	-	&	-	&	-	&	-	&	-	&	-	&	-	&	370.81	\\
10	&	Car2	&	10	&	-	&	-	&	-	&	-	&	-	&	-	&	-	&	-	&	482.50	\\
\hline																							
8	&	Dribble	&	2	&	-	&	0.23	&	0.42	&	-	&	-	&	-	&	1.75	&	0.90	&	1.08	\\
12	&	Dribble	&	3	&	-	&	0.36	&	0.36	&	-	&	-	&	-	&	2.62	&	1.78	&	1.80	\\
16	&	Dribble	&	4	&	-	&	0.51	&	0.51	&	-	&	-	&	-	&	6.94	&	3.23	&	3.14	\\
20	&	Dribble	&	5	&	-	&	0.71	&	0.72	&	-	&	-	&	-	&	10.10	&	4.91	&	4.78	\\
24	&	Dribble	&	6	&	-	&	0.92	&	0.93	&	-	&	-	&	-	&	16.85	&	7.16	&	7.04	\\
28	&	Dribble	&	7	&	-	&	1.08	&	1.09	&	-	&	-	&	-	&	256.84	&	9.48	&	9.70	\\
32	&	Dribble	&	8	&	-	&	1.64	&	1.73	&	-	&	-	&	-	&	84.88	&	12.90	&	13.39	\\
36	&	Dribble	&	9	&	-	&	2.08	&	2.07	&	-	&	-	&	-	&	134.01	&	17.25	&	18.02	\\
40	&	Dribble	&	10	&	-	&	2.74	&	2.76	&	-	&	-	&	-	&	135.79	&	23.07	&	23.49	\\
\hline
\end{tabular}
 \caption{\label{tab:internal-nonlinear}Runtime (s) on nonlinear instances. ``-'' indicates a timeout.}
\end{table*}

We also compare \hnsolve\ with \dreal\ and existing planners,  including
SpaceEx \cite{frehse_spaceexb:_2011,DBLP:conf/hvc/BogomolovFGGPPS14,bogomolov-et-al-sttt2015,bogomolov-etal:cav2012}, CoLin
\cite{coles2012colin}, and UPMurphi \cite{della2009upmurphi}.  
We reproduce previously published results \cite{DBLP:conf/aaai/BogomolovMPW14} for the
SpaceEx, CoLin, and UPMurphi approaches, but report runtimes for
\hnsolve\ and \dreal\ from the same machine, a
2.6 GHz Intel Core i7 and 8GB RAM.  

Our approach inherits some of the limitations of using SpaceEx with
the Bogomolov \etal \cite{DBLP:conf/aaai/BogomolovMPW14} network
representation of PDDL+.  The encoding does not respect the ``must''
semantics of PDDL+ wherein processes and events must occur when
enabled.  However, this limitation is not realized in our chosen
benchmarks because any use of a process or event is advantageous to
the plan.  We also note that dReal (and \hnsolve\ as a result) find
$\delta$-satisfiable solutions to the $\lrf$ encoding.  Owing to the
undecidable nature of nonlinear hybrid systems, dReal cannot guarantee
that a $\delta$-satisfiable solution, which bounds the values of the
continuous variables, contains a realizable plan.  Defining an
appropriately small value for $\delta$ minimizes this concern.  We
also note that dReal in itself is not a full planner.  We report
results for the minimum step length required to find a plan.  A number
of strategies for exploring different step lengths in parallel or in
sequence have been studied in SAT based planning and can be applied
here.  We note that these considerations must be incorporated when
comparing the results for our approach with that of the other planners.

\und{Domains}  We use the Generator and Car
domains from the literature \cite{DBLP:conf/aaai/BogomolovMPW14} and 
the Dribble domain \cite{DBLP:conf/aaai/BryceGMG15}.  We compare on
linear and nonlinear versions of Generator and Car, but only a
nonlinear version of Dribble.

The Car domain includes only atomic actions and processes.  The
actions are to start or stop the Car, and accelerate or decelerate.  The
moving process models one-dimensional kinematics (distance as a
function of velocity and velocity as a function of acceleration) and
the wind-resistance process models the drag effect upon velocity.
Additional actions for acceleration or deceleration
increase the 
branching factor of the problem.  As the problems scale, each instance
$i$ includes actions to accelerate and decelerate by $1, \ldots, i$ units.
The linear and nonlinear versions of the domain differ in whether they include the nonlinear wind-resistance process.

The
Generator domain includes two durative actions: generate, and refuel.
The generate action has a duration of 1000 time units and consumes
fuel at a linear rate.  Its at-end effect satisfies the goal.  Its
overall condition requires that the fuel level is non-negative.  The
instances scale in the number of tanks required to refuel the
Generator so that its overall condition is satisfied.  The refuel
actions increase the fuel level in the Generator continuously, by a
linear rate (in the linear version) or a nonlinear rate (in the
nonlinear version).
For example, the refuel action defines the effects 
linearly as 
\begin{verbatim}
(increase (fuel ?g) (* #t 2))
\end{verbatim}
or nonlinearly
\begin{verbatim}
(increase (ptime ?t) (* #t 1)) 
(increase (fuel ?g) 
          (* #t (* 0.1 (* (ptime ?t)                           
                          (ptime ?t)))))
\end{verbatim}

The Dribble domain involves a process that effects the position $x$ of
a ball.  The position changes continuously based upon the ball
velocity $v$.  The velocity changes continuously due to gravity ($-g$) and drag ($-0.1 v^2$).  The
available actions are dribble($f$) which decrease velocity by $f \in
\{0, 1, 2, 4\}$. The dribble actions have the precondition that velocity is
zero (i.e., the ball is at the top of its arc).  The bounce event
assigns velocity to
$-0.9v$ and has the condition that the ball
position $x$ is zero.  The initial state places the ball at $x=1$ with
velocity $v=0$ and the goal is to  reach $1.5 \leq x \leq
3.0$. The problem, while it does not scale, can be solved for 
plan lengths greater than one.  We find plans (using two-step locking) for steps $k = 8, 12,
16, \ldots, 40$, which correspond respectively to $2, 3, 4, \ldots,
10$ dribble actions interleaved with the same number of bounce events.

\und{Results}  Tables \ref{tab:internal-linear} and
\ref{tab:internal-nonlinear} list runtime results for \dreal\ and
\hnsolve\ on the respective
linear and nonlinear instances.  The columns list the number of
encoding steps $k$, domain, instance, and run times in seconds for
each encoding and solver configuration.  The first three columns of
results are denoted by ``F'' for a hand-coded flat encoding based upon
the instances studied by Bryce \etal \cite{DBLP:conf/aaai/BryceGMG15}.  The second three columns of results
are denoted by ``C'' for the automatically generated parallel
composition of the network encoding into a single automaton.  The last
three columns denoted by ``N'' are the instances encoded with the
network encoding.  Within each group of columns, we denote by ``F'',
``C'', and ``N'' the results that do not use the new \hnsolve\ layer.  The columns with ``+H''
denote results for \hnsolve\ without clause learning, and those with
``+H+L'', for \hnsolve with clause learning.  Entries with ``-''
indicate a timeout of 20 minutes was reached.

\begin{table*}[t]
\centering
\begin{tabular}{|@{ }l@{ }|@{ }l@{ }|@{ }r@{ }|@{ }r@{ }|@{ }r@{ }|@{ }r@{ }|@{ }r@{ }|@{ }r@{ }|@{ }r@{ }|@{ }r@{ }|@{ }r@{ }|@{ }r@{ }|}
\hline
Dom & Planner & 1 & 2 & 3 & 4 & 5& 6 & 7 & 8 \\
\hline
Gen	&	\hnsolve\	&		0.77	&		3.62	&		12.77	&	36.05	&	92.26	&	360.63	&	482.03	&	-	\\
Gen	&	dReal2	&	3.07	&	15.6	&	134.71	&	1699.87	&	-	&	-	&	-	&	-	\\
Gen	&	SpaceEx	&	0.01	&	0.03	&	0.07	&	0.1	&	0.19	&	0.28	&	0.45	&	0.65		\\
Gen	&	CoLin	&	0.01	&	0.09	&	0.2	&	2.52	&	32.62	&	600.58	&	-	&	-	\\
Gen	&	UPMur	&	0.2	&	18.2	&	402.34	&	-	&	-	&	-	&	-	&	-	\\ \hline
Car	&	\hnsolve\	&	1.32	&		1.05	&	1.59	&	2.43	&	3.95	&	6.11	&	9.77	&	18.33	\\
Car	&	dReal2	&	1.07	&	1.17	&	1.16	&	1.22	&	1.23	&	1.29	&	1.26	&	1.21		\\
Car	&	SpaceEx	&	0.01	&	0.01	&	0.01	&	0.03	&	0.04	&	0.05	&	0.06	&	0.07		\\
Car	&	CoLin	&	x	&	x	&	x	&	x	&	x	&	x	&	x	&	x		\\
Car	&	UPMur	&	28.44	&	386.5	&	-	&	-	&	-	&	-	&	-	&	-\\
\hline
\end{tabular}
 \caption{\label{tab:ext} Runtime results (s) on linear Generator and
   Car. ``-'' indicates a timeout.}
\end{table*}

\begin{figure}[t]
\centering \includegraphics[width=.9\linewidth]{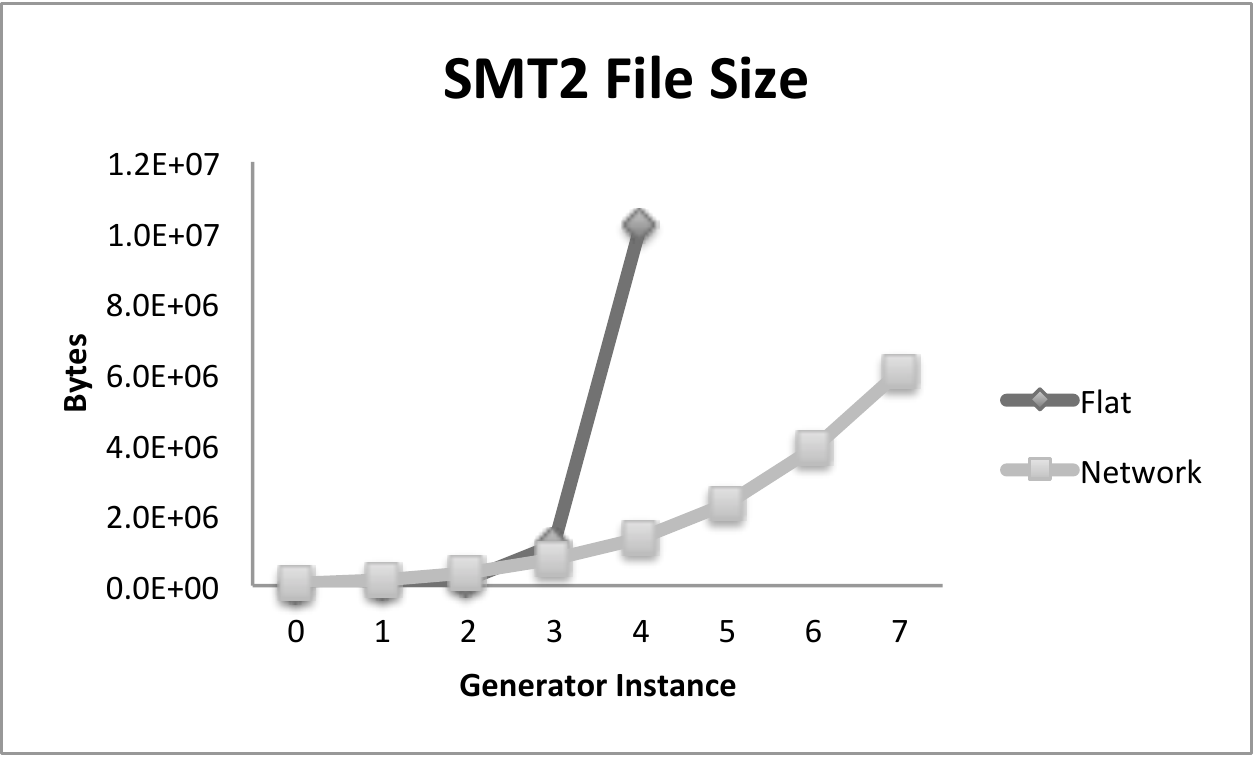}
\caption{\label{fig:size} Generator encoding file size.}
\end{figure}

The results in the Generator domain are listed in the tables with the
number of encoding steps $k$ for the two-step lock encoding used in
the C and N columns, and the steps in parentheses for the F
encodings.  The F encodings model the generate action with three
steps, and each refuel action with two steps.  The C and N encodings
model each action with four steps, but is able to achieve the goal
before releasing the final lock.  Thus, each C and N instance uses one
generate action (3 steps), and a number of refuel actions equal to the
instance number (4 steps each).  The linear results show that all solver
configurations have difficulty scaling on the  F and C encodings, as
reported by \cite{DBLP:conf/aaai/BryceGMG15}.  The critical factor is that
the encoding grows very large with the size of the instances (see
Figure \ref{fig:size}).  Despite
the poor scalability due to the size of the encoding, \hnsolve\ (C+H, C+H+L) can
provide some modest improvement over \dreal\ (C).  The network encoding N performs
significantly better because it uses a tighter encoding.  We also see
the same trends as for the F and C encodings when comparing the different
solver options; \hnsolve\ (N+H, N+H+L) outperforms dReal (N) considerably and clause
learning (+L) has a large impact.

The Car1 instances use the one-step lock and the Car2 instances use
the two-step lock encoding.  The F+H results on Car1 are most similar
to the results reported by \cite{DBLP:conf/aaai/BryceGMG15}.   Our results are somewhat
different because they are based upon dReal3, where the prior work
results were collected on dReal2.  The major difference between these
versions of dReal are its use of the IBEX interval constraint solver
(dReal3), and the Realpaver solver (dReal2).   We see that the Car1
instances heavily favor the F encodings, and the use of \dreal\ over
\hnsolve.  In all caes, we see an improvement over dReal (F, C, N) by
using \hnsolve (+H).  We see less of an improvement, and sometimes
worse performance when using clause learning (+L).  This may be due to
the overhead associated with storing the clauses or a
reshaping of the search space that leads to more backtracking.

The Car2 instances favor the N encoding and
\hnsolve\ with clause learning (N+H+L).  It appears that the difference
is that the relatively shorter encoding lengths in Car1 do not impact
the encoding size as in Car2.  In Car2
where the encoding length is double that of Car1, the \hnsolve\ is
needed to explore the search space.  This result is largely consistent
with the trend demonstrated in the Generator domain, where \hnsolve
performs best as the number of encoding steps increases.

Lastly, the Dribble domain highlights how both \hnsolve\ and the
network encoding have a positive impact upon peformance.  The
F+H configuration and encoding is closest to that reported by Bryce
\etal \cite{DBLP:conf/aaai/BryceGMG15}, and illustrates an improvement over previously published results that we attribute to a difference in \dreal\
version.

Table \ref{tab:ext} compares \hnsolve\  (N+H+L) with the results
reported by \cite{DBLP:conf/aaai/BryceGMG15} (denoted as \dreal{}2, and
similar to F+H) and
the other planners on linear instances of the Generator and Car
domains.  As above, each instance scales the respective number of tanks to fill the
Generator (where each tank is required) and levels of
acceleration/deceleration.  We see that \hnsolve\ scales much better
than \dreal{} on the Generator domain. 
% Upon closer inspection, the reason that \dreal{}2 scales poorly is that it
% spends considerable time in its ICP branch and prune search after finding a
% feasible mode path.  We note that during constraint propagation,
% \dreal{}2 and \dreal{}3 do not use pruning
% operators that are optimized for linear functions. Both \hnsolve\ \dreal{}
% scales well in the Car domain because its heuristic uses reachability
% information to prune unreachable modes.  

%%% Local Variables:
%%% TeX-master: "main"
%%% End:

\section{Related Work}

While PDDL+~\cite{fox2006modelling} has been an accepted language for planning with continuous
change for nearly a decade, very few planners have been able to
handle its expressivity.   Planners either
assume that all continuous change is linear
\cite{DBLP:journals/ai/ShinD05,DBLP:conf/aips/ColesC14,DBLP:conf/aaai/BogomolovMPW14,coles2012colin}
or handle nonlinear change by discretization
\cite{della2009upmurphi}.  

LP-SAT \cite{DBLP:journals/ai/ShinD05} is very similar in spirit to our work
because it uses a SAT solver to solve Boolean constraints and an LP
solver to solve continuous (linear) constraints.  The nature of the
encodings is somewhat different in that our encoding makes use of the
hybrid system semantics of PDDL+ in a very direct fashion.  LP-SAT
more closely resembles classical planning as SAT encodings.  Unlike
our work, LP-SAT
does not incorporate heuristics.

More recent work \cite{brycehp16,cashmore} has extended the LP-SAT
approach by adapting its encoding for use in contemporary SMT solvers,
including dReal and Z3 \cite{Moura2008}.  Unlike our work on \hnsolve,
these works focus solely on planning and not model checking hybrid
systems.  The advantage of focussing on planning encodings is that it
is easier to implement the ``must'' semantics of PDDL+ and adapt
existing techniques for SAT-based planning. Nevertheless, not all
problems are best phrased as PDDL+, and approaches for reasoning about
hybrid systems are necessary.  

Bogomolov \etal \cite{DBLP:conf/aaai/BogomolovMPW14}
 and
Della Penna \etal \cite{della2009upmurphi}, like our work, make use of
the planning as model checking paradigm.  Unlike our work,
Bogomolov \etal \cite{DBLP:conf/aaai/BogomolovMPW14}  encode a network of linear
hybrid automata and we can handle nonlinear automata.
Bogomolov \etal \cite{DBLP:conf/aaai/BogomolovMPW14} use the SpaceEx model
checker \cite{frehse_spaceexb:_2011}, which performs a symbolic search
over the hybrid automata.

Coles \etal \cite{DBLP:conf/aips/ColesC14}
 and \cite{coles2012colin} approach PDDL+ from the perspective 
of heuristic state space search. Coles \etal  \cite{DBLP:conf/aips/ColesC14}
exploit piecewise linear representations of continuous change to
derive powerful pruning conditions for forward heuristic search.
% Given the current lessons gleaned from the rivalry between state space search and
% satisfiability for classical planning, we anticipate approaches for
% state space search in PDDL+ to also inform SAT or SMT-based PDDL+ planning.

\section{Conclusion}
We have described a new specialization of the \dreal\ SMT solver
called \hnsolve\ and an associated network of hybrid automata
encoding.  The combination of \hnsolve\ and the network encoding helps
find PDDL+ plans as the number of encoding steps increases, especially
with the use of a two-step lock encoding.
 We have shown that the
approach scales up reasoning about PDDL+ planning and
that it is competitive with the state of the art.

% In future work, we plan to develop an automatic translation from PDDL+ planning
% as networks of hybrid automata to our solver.  \hnsolve\ can
% incorporate a number of different heuristics into its path generation
% algorithm to improve scalability.  Networks of hybrid automata
% resemble the domain transition graphs used in many classical planning
% heuristics \cite{DBLP:journals/jacm/HelmertHHN14}, and we will
% investigate their application to path generation.  We also see
% opportunities for incorporating numeric information into path
% generation, similar to PDDL+ planning heuristics \cite{DBLP:conf/aips/ColesC14}.

\ifblind
\else
\paragraph{Acknowledgements} This work was supported under ONR
contract N00014-13-1-0090.
\fi

\clearpage
\bibliography{main}
\bibliographystyle{ecai}

\end{document}